\documentclass[conference]{IEEEtran}
\IEEEoverridecommandlockouts
\usepackage{cite}
\usepackage{amsmath,amssymb,amsfonts}
\usepackage{graphicx}
\usepackage{textcomp}
\usepackage{xcolor}
\usepackage{algorithm}
\usepackage{algpseudocode}
\usepackage{float}
\usepackage{bm}

\usepackage[caption=false,font=footnotesize]{subfig}

\def\BibTeX{{\rm B\kern-.05em{\sc i\kern-.025em b}\kern-.08em
    T\kern-.1667em\lower.7ex\hbox{E}\kern-.125emX}}
\begin{document}

\title{Spacetime Splines for Adaptive Robot Motions}
%Spacetime Spline Encoding for Adaptive Robot Motions

\author{
\IEEEauthorblockN{
Boxuan Zhang\textsuperscript{1},
Ante Mari\'c\textsuperscript{2},
and Sylvain Calinon\textsuperscript{2}
}
\thanks{
\textsuperscript{1}Technical University of Munich (TUM), Munich, Germany.

\textsuperscript{2}Idiap Research Institute, Martigny, Switzerland.
}
}

\maketitle

\begin{abstract}
Movement primitives commonly represent robot motions through basis
functions indexed by time or phase. Inspired by spacetime representations employed in physics,
we propose a spline-based formulation in which spatial geometry and
temporal progression are represented within a common set of continuous variables. The proposed method jointly estimates a shared B\'ezier
curve, while allowing spatial and temporal model complexity to be adapted independently (e.g., demonstration-specific phase modulation). As use case, we consider an alternating optimization process that simultaneously estimates path intervals observed in partial demonstrations. Simulation experiments demonstrate accurate joint
recovery, refinement through modifications in model complexity and continuity, and improved robustness to variance in the demonstration set. Finally, we demonstrate online retiming and synchronization capabilities in a patient-turning task assisted by a robot, evaluated with a physical manikin.
\end{abstract}


\section{Introduction} 
 
To adapt and coordinate robot motions, we need to consider both spatial and temporal information. For example, human-guided demonstrations may produce recordings of different durations that can speed up or slow down in different parts of the motion. The recorded paths and sampling rates may also differ, and some recordings may contain only part of the overall motion, starting and stopping at different points of the full path.

Spline representations provide compact continuous models of robot motion~\cite{miyamoto1996kendama,5580104,bmp,ralli}. B\'ezier curves provide one such representation that allows for easy modification of through geometrically-interpretable control points and favorable mathematical properties. For example, De Casteljau's algorithm can be used as a stable method for evaluating and subdividing these curves~\cite{farin2002curves,LineartimeGA}.
Temporal alignment and phase-based methods account for nonlinear timing differences or separate path geometry from timing~\cite{sakoe1978dtw,trajectorylearningdtw,raket2016separating,braglia2025arclength,braglia2025phaseindependent,rhodes2023timewarpvae}, while others learn from incomplete or partial demonstrations~\cite{sun2019incomplete,pervez2017incomplete,silverio2019uncertainty,li2023partial}. A common thread is that temporal and spatial aspects are encoded and
optimized separately.
 
\begin{figure}[t]
    \centering
    \includegraphics[width=\columnwidth]{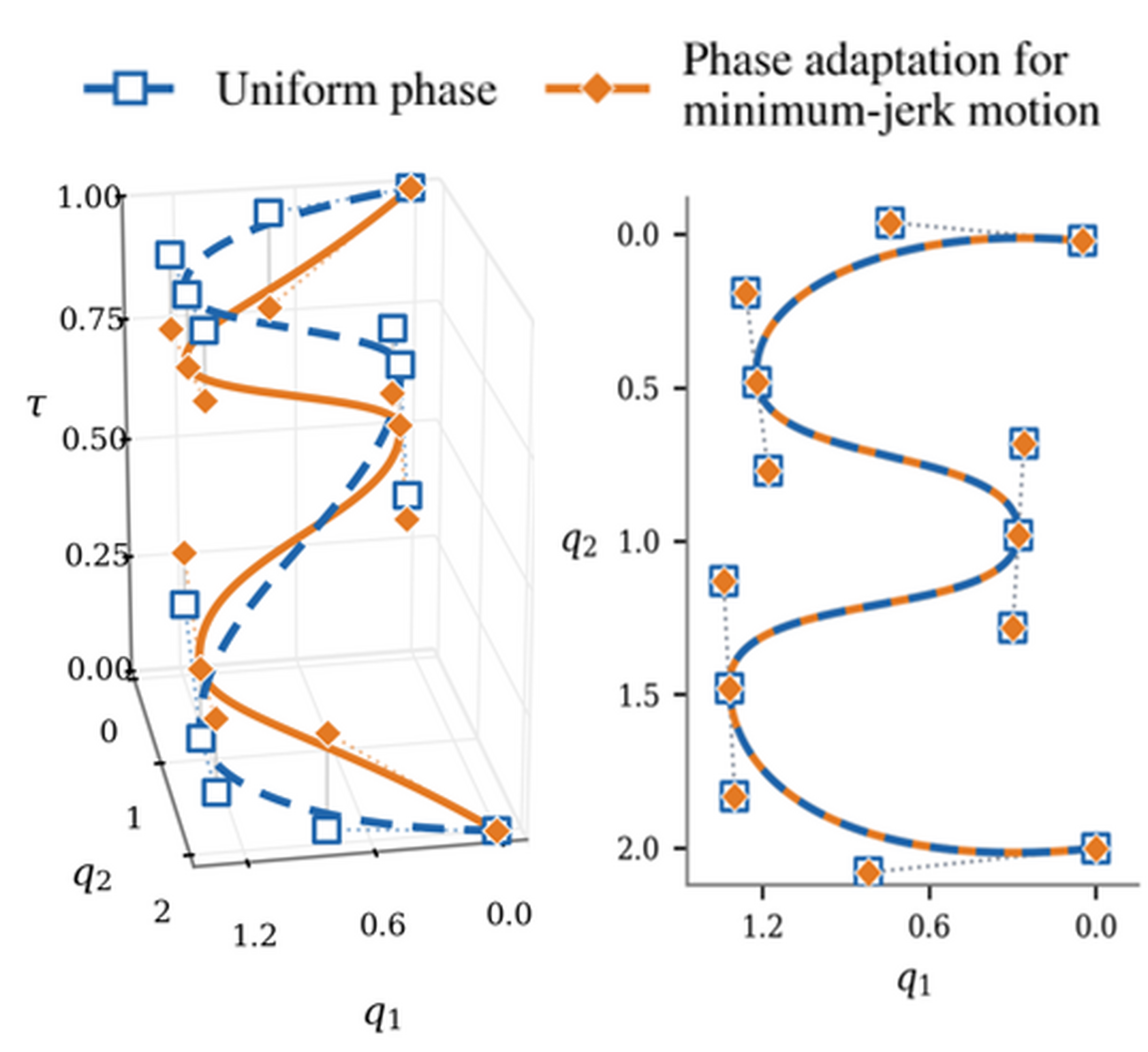}
    \caption{The spacetime spline representation can be used to preserve a shared spatial path geometry while modulating the motion timing by adapting the phase component of the augmented state space. Left: uniform (standard) in $(q_1,q_2,\tau)$ space and phase
adaptation to provide a minimum-jerk motion. Right: Same movements in the
spatial $q_1$--$q_2$ plane, showing that the path geometry is preserved
under phase adaptation.}
    \label{fig:overview}
\end{figure}

Inspired by spacetime representations in physics, we propose a continuous dynamic curve representation that homogeneously encodes spatial geometry and motion timing as an augmented state space. 
%Demonstration coverage is represented by an interval on the shared path. 
Figure~\ref{fig:overview} shows an example in which different phase modulations produce different spacetime curves preserving a spatial path. We consider in this paper a use case in which we jointly estimate a shared geometric path with a demonstration-specific modulation of the phase function that describes progress along that path. 
%We adapt the numbers of geometry and phase segments independently. 
We utilize and alternating optimization process that adapts path geometry by modifying spatial coordinates of the control points,
and motion timing through the temporal coordinates of the control points. This allows us to estimate path intervals in partial demonstrations of a longer motion.

The main contributions of this work are: 
\begin{itemize} 
    \item \textbf{Spacetime dynamic curve representation:} a modular continuous
    spline encoding that homogeneously couples path geometry and phase, while allowing independent spatial and temporal adaptation.
 
    \item \textbf{Coverage estimation in partial demonstrations:} joint inference of path intervals observed in partial demonstrations. 
 
    \item \textbf{Spatial and temporal adaptation:} adaptive refinement of shared B\'ezier geometry and demonstration-specific phase for online retiming and synchronization of robot motions.
\end{itemize}
We evaluate joint recovery, structural preservation, and adaptive optimization in a set of simulation experiments. Finally, we demonstrate collaborative synchronization under changing human timing and temporary interruptions in a patient-turning task assisted by a robot, evaluated with a physical manikin.

\section{Related Work}

\subsection{Spline-Based Motion Representation and Adaptive Fitting}

Spline representations provide compact continuous models of robot
motion~\cite{bmp,ralli}, with B\'ezier and B-spline models offering
smoothness, explicit control-point structure, and analytical
derivatives~\cite{farin2002curves,LineartimeGA}.

B-spline Movement Primitives (BMPs)~\cite{bmp} combine B\nobreakdash-spline representations with movement-primitive parameterizations, while concatenated B\'ezier splines have been used as continuous motion encoding for distance-field and dynamical-system construction~\cite{ralli}. Spline parameters have also been adopted as structured action representations for policy learning~\cite{tianspline,han2026bsplinepolicy}. In parallel, free-knot and adaptive spline methods optimize or refine knot placement according to the data, allowing representation capacity to be allocated nonuniformly~\cite{jupp1978freeknot,miyata2003adaptive,li2005adaptive,park2007adaptive,zhao2011adaptive}.

Our work builds on these ideas but addresses an additional structural ambiguity: reconstruction error may originate from insufficient complexity of the spatial and temporal models representing the robot motion. Including both position and velocity information in an augmented state space enables additional dynamic information to be encoded in the curve. However, this approach has limitations, discussed in ~\cite{tekden26reactive}, which can be addressed through the use of a phase variable. We propose to adapt geometry and phase components of the joint spacetime curve in an alternating optimization scheme and evaluate candidate refinements under a common regularized objective. 

\subsection{Temporal Alignment and Geometry--Phase Modeling}

Temporal variations are common when learning motion from multiple demonstrations. Dynamic time warping (DTW)~\cite{sakoe1978dtw} and related registration methods~\cite{trajectorylearningdtw} are commonly used to align demonstrations with nonlinear timing differences. This warping process is typically used as a pre-processing step to align trajectories and requires the discretization of the curve and the tuning of hyperparameters to define the allowed distortion, which typically depends on the chosen resolution of the curve (number of discretization points) and which often only considers local warping. Our approach instead relies on a continuous time adaptation function that does not require discretization. 

In functional data analysis, temporal misalignment is often treated as phase variation and separated from spatial variation. Curve registration methods estimate nonlinear transformations of the independent variable in order to align salient structures across observations~\cite{ramsay2018registration}. Raket et al.~\cite{raket2016separating} showed that excessive alignment can absorb genuine spatial variability into the estimated warping functions. This motivates joint formulations in which the complexity of the temporal transformation and the spatial representation must be balanced rather than treating alignment as an independent preprocessing step.

Recent robotics methods further separate path geometry from timing. Braglia et al.~\cite{braglia2025arclength} proposed arc-length-based alignment using a time-independent path parameterization. Their subsequent phase-independent Dynamic Movement Primitive (DMP)
formulation further decouples the demonstrated geometric path from its
timing law, allowing execution phase to be redesigned independently for applications including human--robot co-manipulation and time-optimal motion generation~\cite{braglia2025phaseindependent}. TimewarpVAE~\cite{rhodes2023timewarpvae} similarly incorporates differentiable temporal warping into representation learning.

We jointly estimate a shared geometric spline and demonstration-specific monotone phase functions, with the recovered phase remaining an explicit component of the motion representation and allowing for subsequent constraints or modifications. Each demonstration is additionally associated with an explicit coverage interval on the shared path, allowing the encoding of temporal progression and spatial coverage.

\subsection{Learning from Partial Demonstrations}

Some work addresses incomplete supervision, such as missing action labels or partially observed state--action information~\cite{sun2019incomplete}. Other methods consider demonstrations that cover only part of a task. Pervez et al.~\cite{pervez2017incomplete} combined temporal synchronization with movement learning from incomplete teleoperation demonstrations. Silv\'erio et al.~\cite{silverio2019uncertainty} combined movement models learned from partial subtasks through uncertainty\nobreakdash-aware controller fusion. Partial demonstrations have also been used to guide policy learning without requiring exact trajectory reproduction~\cite{li2023partial}.

Our setting differs in that a partial demonstration is modeled as an observation of an unknown contiguous interval of a common continuous motion. Its coverage is estimated jointly with the shared geometry and demonstration-specific phase. Complete and partial demonstrations therefore contribute to a shared representation, while spatial coverage remains separate from temporal progression.

\section{Methodology}
\label{sec:method}

\subsection{Spacetime Motion Encoding}
Given samples $\bm{q}_i(t_{i,n})\in\mathbb{R}^D$,
$n=1,\ldots,N_i$, let \mbox{$\mathcal I=\{1,\ldots,N\}$} be
partitioned into complete and partial demonstration sets
$\mathcal I_{\mathrm{full}}$ and $\mathcal I_{\mathrm{part}}$.
Using normalized recording time $r\in[0,1]$, we encode
motion homogeneously in augmented spacetime as
\begin{align}
    \tau_i(r)
    &=a_i+(b_i-a_i)u_i(r),
    \label{eq:temporal_coordinate}\\
    \bm{z}_i(r)
    &=
    \begin{bmatrix}
        \bm{g}(\tau_i(r))\\
        r
    \end{bmatrix}
    \in\mathbb{R}^{D+1},
    \label{eq:spacetime_curve}
\end{align}
where $\bm{g}:[0,1]\rightarrow\mathbb{R}^D$ is the shared
spatial map and $\tau_i(r)$ describes progression along it.
The monotone profile $u_i:[0,1]\rightarrow[0,1]$ controls
this progression over the coverage interval $[a_i,b_i]$.
The reconstructed trajectory is the spatial projection
\begin{equation}
    \hat{\bm{q}}_i(t)
    =
    \bm{g}\bigl(\tau_i(r_i(t))\bigr),
    \qquad
    r_i(t)
    =
    \frac{t-t_{i,1}}{t_{i,N_i}-t_{i,1}}.
    \label{eq:factorization}
\end{equation}
Complete demonstrations have $(a_i,b_i)=(0,1)$, while partial coverage is estimated jointly with the spatial
and temporal coordinate parameters.
One complete anchor
\begin{equation}
    u_{i_{\mathrm{anc}}}(r)=r,\qquad
    (a_{i_{\mathrm{anc}}},b_{i_{\mathrm{anc}}})=(0,1),
    \label{eq:anchor}
\end{equation}
fixes the common monotone reparameterization ambiguity.

\subsection{Coordinate-Specific Spline Parameterization}
We use the cubic Bernstein basis vector
\begin{equation}
    \bm{\phi}_B(\xi)
    =[B_0^3(\xi),\ldots,B_3^3(\xi)]^{\top},
    \label{eq:bernstein_basis}
\end{equation}
with scalar basis functions $B_\nu^3(\xi)=\binom{3}{\nu}(1-\xi)^{3-\nu}\xi^\nu$, \mbox{$\nu=0,\ldots,3$}.

The shared spatial map $\bm{g}$ uses $M_g$ cubic B\'ezier
segments over knots $\mathcal K_g=\{\kappa_0,\ldots,\kappa_{M_g}\}$,
with \mbox{$0=\kappa_0<\cdots<\kappa_{M_g}=1$}.
For parameter value $s\in[\kappa_m,\kappa_{m+1}]$, the spatial map is expressed as
\begin{equation}
    \bm{g}_m(s)=\bm{P}_m^{\top}\bm{\phi}_B(\xi_m(s)),
    \qquad
    \xi_m(s)=\frac{s-\kappa_m}{\kappa_{m+1}-\kappa_m},
    \label{eq:geometry}
\end{equation}
where $\bm{P}_m=[\bm{p}_{m,0},\ldots,\bm{p}_{m,3}]^{\top}
\in\mathbb{R}^{4\times D}$ represents a matrix of control points $\bm{p}_{m,\nu}\in\mathbb{R}^D$. For demonstration $i$, $\xi_m$ represents the local segment parameter and
phase progression is parametrized as $s=\tau_i(r)$.

Adjacent segments share a control point
$\bm{p}_{m,3}=\bm{p}_{m+1,0}$, ensuring $C^0$ continuity.
At selected $G^1$ junctions, we further impose
\begin{equation}
\begin{aligned}
    \bm{p}_{m,2}&=\bm{p}_{m+1,0}-\ell_{m+1}^-\bm{d}_{m+1},\\
    \bm{p}_{m+1,1}&=\bm{p}_{m+1,0}+\ell_{m+1}^+\bm{d}_{m+1},
    \label{eq:g1}
\end{aligned}
\end{equation}
where $\bm{d}_{m+1}$ is a unit tangent direction at $\bm{p}_{m+1,0}$ and
$\ell_{m+1}^-,\ell_{m+1}^+>0$ are handle lengths.
This aligns tangent directions while allowing different
magnitudes.

Analogously to the above, non-anchor temporal coordinates use a piecewise-cubic profile $u_i$ over $M_i$ segments
\begin{equation}
    u_i(r)=\bm{c}_{i,k}^{\top}\bm{\phi}_B(v_{i,k}(r)),
    \qquad
    v_{i,k}(r)=
    \frac{r-\eta_{i,k}}{\eta_{i,k+1}-\eta_{i,k}},
    \label{eq:phase}
\end{equation}
with control points $\bm{c}_{i,k}=[c_{i,k,0},\ldots,c_{i,k,3}]^{\top}$, spline parameter $r$ and local segment parameter $v_{i,k}$. In the above, $\eta_{i,k}$ represents the $k$-th knot of demonstration $i$.
Monotonicity is enforced by
\begin{equation}
    c_{i,k,0}\leq c_{i,k,1}\leq
    c_{i,k,2}\leq c_{i,k,3},
    \label{eq:phase_mono}
\end{equation}
together with $u_i(0)=0$, $u_i(1)=1$, and shared endpoints
$c_{i,k,3}=c_{i,k+1,0}$ between adjacent segments.
These constraints prevent temporal reversal while allowing
pause-like progression and ensuring $C^0$ continuity. Further continuity constraints can be imposed by matching derivatives at internal knots.

\subsection{Alternating Spacetime Optimization}
For partial demonstrations, coverage is parameterized by
$\alpha_i,\beta_i\in\mathbb{R}$.
With $\sigma(x)=(1+\exp(-x))^{-1}$ and minimum coverage
length $\Delta_{\min}\in(0,1)$,
% \begin{align}
%     a_i &=(1-\Delta_{\min})\sigma(\alpha_i),
%     \nonumber\\
%     b_i &=a_i+\Delta_{\min}
%     +(1-a_i-\Delta_{\min})\sigma(\beta_i),
%     \label{eq:coverage}
% \end{align}
\begin{equation}
    a_i =(1-\Delta_{\min})\sigma(\alpha_i),\;
    b_i =a_i+\Delta_{\min}
    +(1-a_i-\Delta_{\min})\sigma(\beta_i),
    \label{eq:coverage}
\end{equation}
ensuring $0\leq a_i<b_i\leq1$.

Let $\bm{\theta}_g$ and $\bm{\theta}_{u,i}$ collect the
spatial and temporal profile parameters, respectively.
For each non-anchor demonstration, we define
\begin{equation}
    \bm{\theta}_i=
    \begin{cases}
        \bm{\theta}_{u,i},
        &i\in\mathcal I_{\mathrm{full}},\\
        [\bm{\theta}_{u,i}^{\top},\alpha_i,\beta_i]^{\top},
        &i\in\mathcal I_{\mathrm{part}},
    \end{cases}
    \label{eq:local_variables}
\end{equation}
and the full parameter vector $\bm{\theta}=\operatorname{col}\bigl(\bm{\theta}_g, \{\bm{\theta}_i\}_{i\neq i_{\mathrm{anc}}}\bigr)$,
where $\operatorname{col}(\cdot,\cdot)$ stacks blocks vertically.
Using the mean squared reconstruction error
\begin{equation}
    e_i=\frac{1}{N_i}\sum_{n=1}^{N_i}
    \|\bm{q}_i(t_{i,n})-\hat{\bm{q}}_i(t_{i,n})\|_2^2,
    \label{eq:reconstruction}
\end{equation}
for a fixed spline structure, we minimize
% \begin{equation}
%     \begin{split}
%         \mathcal L(\bm{\theta})
%         ={}&\frac{1}{N}\sum_{i=1}^{N}e_i+\lambda_g r_g
%         +\lambda_s\sum_{i\neq i_{\mathrm{anc}}}r_{s,i}\\
%         &+\lambda_{\mathrm{cov}}
%         \sum_{i\in\mathcal I_{\mathrm{part}}}[1-(b_i-a_i)].
%     \end{split}
%     \label{eq:objective}
% \end{equation}
\begin{equation}
    \mathcal L(\bm{\theta})
        =\frac{1}{N}\sum_{i=1}^{N}e_i+\lambda_g r_g
        +\lambda_s \!\!\sum_{i\neq i_{\mathrm{anc}}}\!\!\! r_{s,i}
        +\lambda_{\mathrm{cov}}
        \!\!\sum_{i\in\mathcal I_{\mathrm{part}}}\!\!\!\! \big(1-(b_i-a_i)\big).
    \label{eq:objective}
\end{equation}
Curvature and coverage intervals are regularized using
\begin{equation}
    r_g=\int_0^1\|\bm{g}''(s)\|_2^2\,ds,\qquad
    r_{s,i}=\int_0^1|u_i''(r)|^2\,dr,
    \label{eq:regularizers}
\end{equation}
with second derivatives evaluated segmentwise.

Block-coordinate optimization first fixes the spatial block
and updates non-anchor temporal blocks in parallel as
\begin{equation}
    \begin{aligned}
        \bm{\theta}_i^+
        &=\arg\min_{\bm{\theta}_i\in\mathcal F_i}
        \Bigl[\frac{1}{N}e_i+\lambda_s r_{s,i}
        +\lambda_{\mathrm{cov}}(1-b_i+a_i)\Bigr].
    \end{aligned}
    \label{eq:parallel_phase}
\end{equation}
Here, $\mathcal F_i$ enforces monotonicity, endpoints,
coverage, and the selected continuity constraints.
%The coverage penalty vanishes for complete demonstrations.
Fixing these temporal blocks, we then update
\begin{equation}
    \bm{\theta}_g^+
    =\arg\min_{\bm{\theta}_g}
    \left[\frac{1}{N}\sum_{i=1}^{N}e_i+\lambda_g r_g\right],
    \label{eq:geometry_update}
\end{equation}
using regularized linear least squares for all-$C^0$
junctions and constrained nonlinear least squares with $G^1$
junctions.
Updates alternate until the relative change of $\mathcal L$
falls below a threshold.

\begin{table*}[!t]
\centering
\caption{Recovery on complete and partial demonstrations
(mean $\pm$ standard deviation over 30 runs). Lower is better.
DMP and ProMP are evaluated on complete demonstrations only because their
standard formulations do not infer unknown coverage intervals.}
\label{tab:q1_recovery}
\footnotesize
\renewcommand{\arraystretch}{1.2}
\setlength{\tabcolsep}{6.0pt}
\begin{tabular*}{\textwidth}{@{\extracolsep{\fill}}llcccc@{}}
\hline
\textbf{Setting} & \textbf{Method}
& \textbf{Traj. RMSE}
& \textbf{Geom. RMSE}
& \textbf{Phase RMSE}
& \textbf{Cov. MAE} \\
\hline

Complete
& Ours
& $0.0154\pm0.0012$
& $0.0127\pm0.0014$
& $0.0170\pm0.0004$
& -- \\

& DTW+spline
& $\mathbf{0.0105\pm0.0017}$
& $\mathbf{0.0061\pm0.0031}$
& $\mathbf{0.0055\pm0.0003}$
& -- \\

& Linear phase
& $0.1368\pm0.0134$
& $0.0523\pm0.0227$
& $0.1728\pm0.0000$
& -- \\

& DMP
& $0.1372\pm0.0136$
& $0.0550\pm0.0202$
& -- & -- \\

& ProMP
& $0.1367\pm0.0136$
& $0.0526\pm0.0227$
& -- & -- \\

\hline

Partial
& Ours
& $0.0200\pm0.0012$
& $0.0187\pm0.0032$
& $0.0270\pm0.0031$
& $0.0348\pm0.0047$ \\

& DTW+spline
& $\mathbf{0.0110\pm0.0020}$
& $\mathbf{0.0060\pm0.0032}$
& $\mathbf{0.0057\pm0.0003}$
& $\mathbf{0.0057\pm0.0013}$ \\

& Linear phase
& $0.0753\pm0.0046$
& $0.0992\pm0.0079$
& $0.1447\pm0.0027$
& $0.1379\pm0.0046$ \\

\hline
\end{tabular*}
\end{table*}

\subsection{Adaptive Coordinate Refinement}

Starting from few segments per coordinate block,
we accumulate reconstruction errors over the optimized spatial and temporal partitions.
Subsequently, we use De Casteljau subdivision to initialize candidate splits on segments with the largest mean errors while retaining encoded curves.

Spatial junctions are additionally tested for $G^1$ continuity.
Namely, for junction $j=m+1$ with nonzero handles
$\bm{h}_j^-=\bm{p}_j-\bm{p}_{m,2}$ and
$\bm{h}_j^+=\bm{p}_{m+1,1}-\bm{p}_j$, we define
% \begin{equation}
%     \begin{aligned}
%         \vartheta_j
%         &=
%         \arccos
%         \frac{(\bm{h}_j^-)^{\top}\bm{h}_j^+}
%              {\|\bm{h}_j^-\|_2\|\bm{h}_j^+\|_2},\\
%         \rho_j
%         &=
%         \frac{\max(\|\bm{h}_j^-\|_2,\|\bm{h}_j^+\|_2)}
%              {\min(\|\bm{h}_j^-\|_2,\|\bm{h}_j^+\|_2)}.
%     \end{aligned}
%     \label{eq:g1_candidates}
% \end{equation}
\begin{equation}
        \vartheta_j
        =
        \arccos
        \frac{(\bm{h}_j^-)^{\top}\bm{h}_j^+}
             {\|\bm{h}_j^-\|_2\|\bm{h}_j^+\|_2},\quad
        \rho_j
        =
        \frac{\max(\|\bm{h}_j^-\|_2,\|\bm{h}_j^+\|_2)}
             {\min(\|\bm{h}_j^-\|_2,\|\bm{h}_j^+\|_2)}.
    \label{eq:g1_candidates}
\end{equation}

For thresholds $\vartheta_{\max}\in[0,\pi]$ and
$\rho_{\max}\geq1$, junctions satisfying
$\vartheta_j\leq\vartheta_{\max}$ and
$\rho_j\leq\rho_{\max}$ are reoptimized under the
spatial $G^1$ parameterization and retained if the increase in~\eqref{eq:objective} remains within a threshold.
% \begin{equation}
%     \delta_j^{G^1}
%     =
%     \frac{L_j^{G^1}-L_{\mathrm{cur}}}
%          {\max(L_{\mathrm{cur}},\epsilon_0)}
%     \leq\epsilon_{G^1},
%     \label{eq:g1_acceptance}
% \end{equation}
% where $L_{\mathrm{cur}}$ and $L_j^{G^1}$ are the values
% of $\mathcal L$ before and after reoptimization with the
% spatial $G^1$ constraint at junction $j$.
% Here, $\epsilon_{G^1}\geq0$ bounds the relative objective
% increase, and $\epsilon_0>0$ is a numerical constant.

For a coordinate-specific structure
\mbox{$\mathcal M=(\mathcal K_g,\mathcal C,
\{\mathcal K_{u,i}\}_{i=1}^{N})$},
where $\mathcal C$ specifies the $C^0$ or $G^1$
continuity class at each spatial junction,
subdivision candidates are compared using
\begin{equation}
    \mathcal J(\mathcal M)
    =
    \min_{\bm{\theta}}
    \mathcal L(\bm{\theta}\mid\mathcal M)
    +\gamma_g M_g
    +\gamma_s
    \sum_{i\neq i_{\mathrm{anc}}}M_i,
\end{equation}
where $\gamma_g,\gamma_s>0$ penalize spatial and
temporal segments, and alternating optimization
approximates the inner minimum. A candidate structure $\mathcal M'$ is accepted only if
\mbox{$\mathcal J(\mathcal M') < \mathcal J(\mathcal M)-\epsilon_{\mathrm{struct}}$},
where $\epsilon_{\mathrm{struct}}\geq0$ is the minimum
required score reduction.
If both split candidates are admissible, the one with the
larger score reduction is selected.
Refinement stops when neither split is accepted or the
target reconstruction error is reached.

After optimization, the spatial projection is evaluated
using a monotone temporal coordinate $\tau_d(t)$:
\begin{equation}
    \bm{q}_d(t)=\bm{g}(\tau_d(t)),\qquad
    \dot{\bm{q}}_d(t)
    =\bm{g}'(\tau_d(t))\dot{\tau}_d(t).
    \label{eq:execution_mapping}
\end{equation}
The velocity relation holds wherever $\bm{g}$ is
differentiable.
Linear profiles spanning $[0,1]$ or $[a_i,b_i]$ give
nominal or partial traversal, respectively, and nonlinear profiles permit retiming and pause-like progression.
As a use case example, we demonstrate online adaptation of $\tau_d(t)$ in an assistive task experiment in Sec.~\ref{sec:real_robot}.

\section{Experiments}

\subsection{Simulation Experiments}

We evaluate three aspects of the proposed formulation. Q1: joint recovery of geometry, phase, and partial coverage; Q2: preservation of geometric structure; and Q3: the role of alternating optimization and adaptive model complexity.

\textit{Joint recovery and robustness (Q1).}
We evaluate whether the proposed formulation can jointly recover a shared
geometric path, demonstration-specific phase evolution, and partial coverage.
We use six representative shapes from the LASA handwriting
dataset~\cite{khansari2011learning}: Angle, CShape, Sine, Line,
JShape, and Trapezoid as latent paths $\bm{g}^{\star}$. For each shape, seven
demonstrations are generated with nonlinear timing and observation noise, while
partial trials additionally expose only a sub-interval of the latent path.
Five trials are performed per shape, resulting in 30 runs for each evaluation
setting. One complete demonstration is fixed as the canonical anchor.
Partial coverage is constrained to a minimum span of $0.12$, and alternating
optimization is run for at most 30 iterations with a convergence tolerance
of $10^{-7}$.

\begin{figure}[t]
    \centering
    \includegraphics[width=\columnwidth, height=130px]{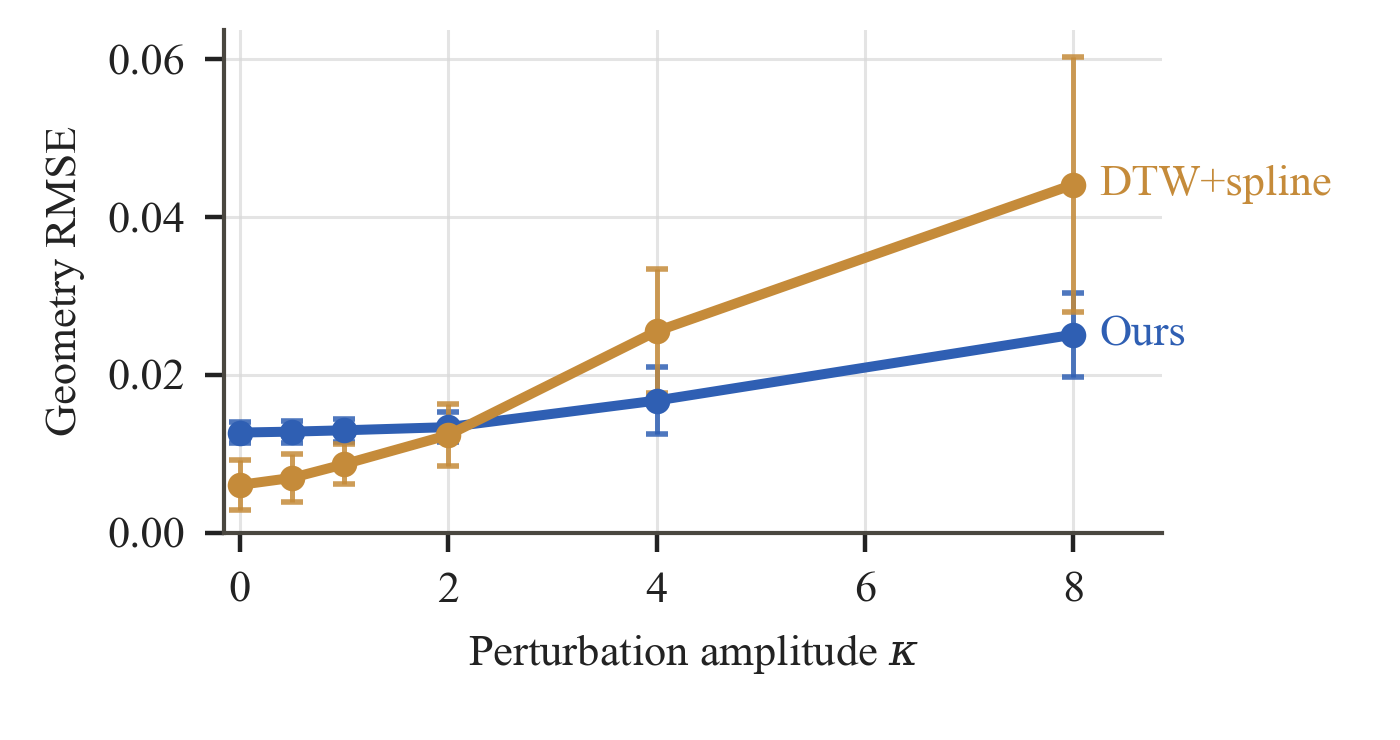}
    \caption{Geometry recovery under increasing independent spatial perturbations
    across demonstrations. The horizontal axis shows the perturbation amplitude
    $\kappa$ in units of $10^{-2}$. Without a privileged clean reference, our method
    degrades more slowly than DTW+spline as the perturbation amplitude increases.}
    \label{fig:q1_robustness}
\end{figure}

We compare our method with DTW+spline and a fixed linear-phase model.
Dynamic Movement Primitives (DMPs)~\cite{dmp} and Probabilistic Movement
Primitives (ProMPs)~\cite{promp} are additionally included for complete
demonstrations. For DTW+spline, demonstrations are aligned to a selected
reference before spline fitting; for partial demonstrations, the aligned
endpoints determine the estimated coverage interval. We report trajectory
and geometry root\nobreakdash-mean\nobreakdash-square error (RMSE), phase RMSE where applicable,
and coverage endpoint mean absolute error (MAE) for partial demonstrations.

Table~\ref{tab:q1_recovery} evaluates the clean-reference regime, which
deliberately favors reference-based alignment. DTW+spline performs best in this
setting, as expected from its access to a reliable reference trajectory.

We next remove this privileged reference by independently
perturbing every demonstration with a smooth spatial deviation
$\bm{\delta}_i$ scaled by amplitude $\kappa$,
\begin{equation}
\bm{q}_i(t)
=
\bm{g}^{\star}(s_i(t))
+\bm{\delta}_i(s_i(t))
+\bm{\epsilon}_i(t),
\label{eq:multidemo_perturb}
\end{equation}
where $\bm{\epsilon}_i(t)$ denotes Gaussian observation noise
and $\bm{\delta}_i$ is generated independently for each
demonstration and centered across the ensemble, such that
$\bm{g}^{\star}$ remains the latent shared path.
Thus, no individual demonstration provides an unperturbed spatial reference.

As shown in Fig.~\ref{fig:q1_robustness}, DTW+spline is more accurate at low perturbation levels, whereas our method degrades more slowly as
inter-demonstration spatial variation increases. This shows that reference-based alignment is effective when a reliable
template is available, while joint estimation is more robust when the shared geometry must be inferred from mutually inconsistent demonstrations.

\textit{Structural preservation and adaptive continuity (Q2).}
\begin{figure}[t]
    \centering
    \subfloat[]{%
        \includegraphics[width=0.486\columnwidth]
        {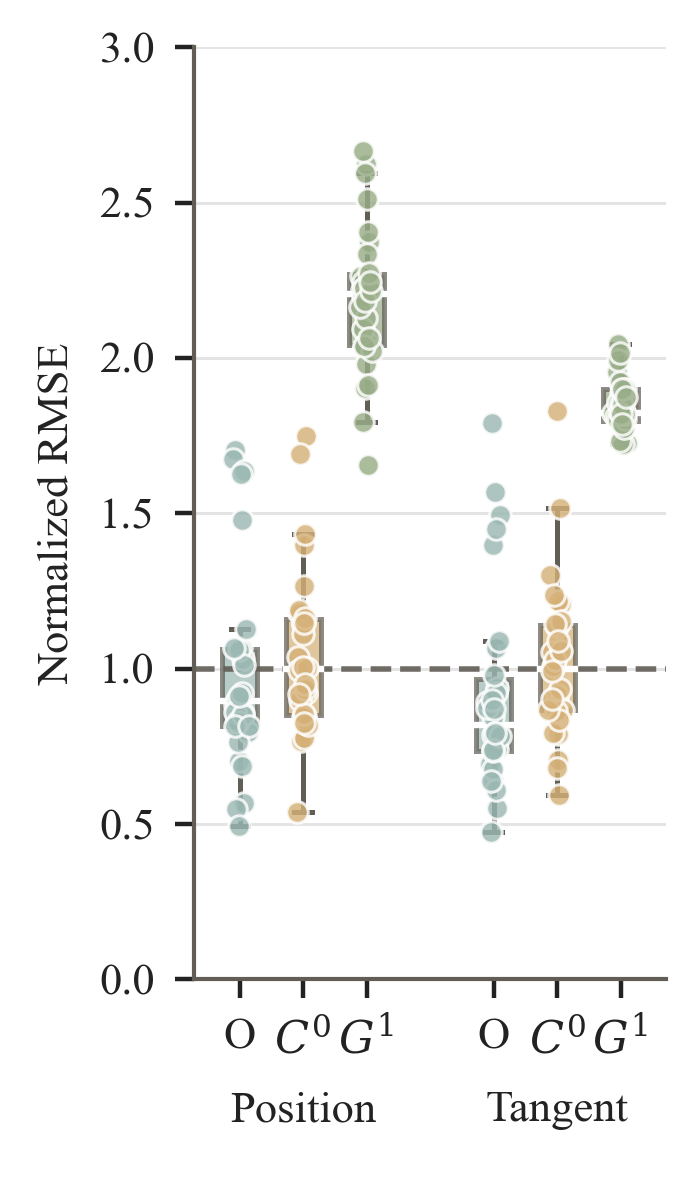}%
        \label{fig:q2_structure}%
    }\hfill%
    \subfloat[]{%
        \includegraphics[width=0.500\columnwidth]
        {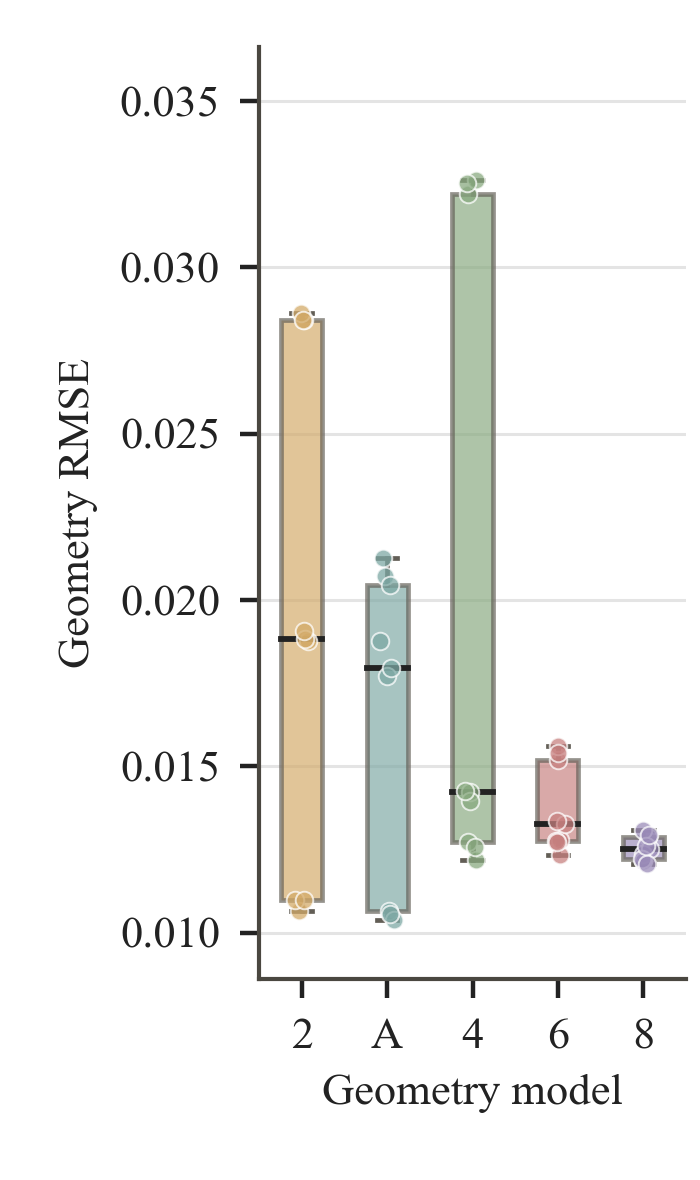}%
        \label{fig:adaptive_refinement}%
    }

    \caption{(a) Structural ablation.
    Position and tangent RMSEs are normalized by the
    corresponding medians of the all-$C^0$ variant.
    $O$, $C^0$, and $G^1$ denote our method, all-$C^0$,
    and all-$G^1$, respectively.
    (b) Adaptive geometry complexity.
    $A$ denotes the adaptive model, which selects
    $3.89 \pm 1.45$ segments (mean $\pm$ standard deviation);
    the other labels indicate fixed segment counts.
    Mean runtimes for $2$, $A$, $4$, $6$, and $8$ are
    $0.58$, $21.0$, $18.7$, $121$, and $689$\,s, respectively.}
    \label{fig:structure_and_ref}
\end{figure}
We evaluate continuity selection on paths containing both smooth, continuous transitions and discontinuities. Over 30 runs, each trial contains five
demonstrations of 50 samples with Gaussian observation noise
($\sigma=0.009$). For experiments involving temporal-profile estimation,
we use $C^2$ continuity. Here, ground-truth phase is supplied to all
variants in order to evaluate geometric continuity in isolation.

Figure~\ref{fig:q2_structure} shows that the adaptive formulation
achieves the lowest error in both metrics. This indicates that selecting continuity
locally avoids the loss of tangent consistency of All-$C^0$ while
preventing the over-smoothing introduced by globally enforced $G^1$
continuity.

\begin{table}[!t]
\centering
\caption{Ablation of the joint optimization
(mean $\pm$ standard deviation over 15 runs). Lower is better.}
\label{tab:q3_ablation}
\footnotesize
\renewcommand{\arraystretch}{1.2}
\setlength{\tabcolsep}{3.0pt}
\resizebox{\columnwidth}{!}{%
\begin{tabular}{@{}lccc@{}}
\hline
\textbf{Method}
& \textbf{Geom. RMSE}
& \textbf{Phase RMSE}
& \textbf{Cov. MAE} \\
\hline

Ours
& $\mathbf{0.0210\pm0.0017}$
& $\mathbf{0.0282\pm0.0027}$
& $\mathbf{0.0347\pm0.0049}$ \\

One-shot
& $0.0541\pm0.0019$
& $0.0687\pm0.0017$
& $0.0813\pm0.0083$ \\

Fixed coverage
& $0.0571\pm0.0075$
& $0.1036\pm0.0013$
& $0.2070\pm0.0000$ \\

Linear phase
& $0.0945\pm0.0100$
& $0.1431\pm0.0042$
& $0.1359\pm0.0076$ \\

\hline
\end{tabular}%
}
\end{table}

\textit{Alternating optimization and adaptive complexity (Q3).}
Table~II shows that removing alternating updates or fixing coverage or
phase substantially degrades joint recovery, confirming the importance of
optimizing geometry, phase, and coverage together. Fig.~\ref{fig:structure_and_ref}\subref{fig:adaptive_refinement} compares the automatically selected adaptive
complexity with fixed-complexity alternatives, illustrating the corresponding
accuracy--runtime trade-off.

% \begin{figure}[t]
%     \centering
%     \includegraphics[width=0.86\columnwidth]{fig7_adaptive_refinement.pdf}
%     \caption{Adaptive geometric complexity compared with fixed-complexity models.
% The adaptive procedure automatically selects model complexity, while fixed
% models illustrate the corresponding accuracy--runtime trade-off.}
%     \label{fig:adaptive_refinement}
% \end{figure}

The implementation also benefits substantially from analytic phase
derivatives and different demonstrations can be processed in parallel. All timing measurements were obtained on an Intel Core i9 CPU. For 32 demonstrations, median fitting time decreases from $88.5$\,s with serial finite differences to $5.81$\,s with analytic gradients and $3.04$\,s with parallel
demonstration updates. A per-demonstration update
requires $16.9$\,ms median computation time and $29.8$\,ms at the
95th percentile.

% \begin{table}[t]
% \centering
% \caption{Ablation of the joint optimization
% (mean $\pm$ standard deviation over 15 runs). Lower is better.}
% \label{tab:q3_ablation}
% \scriptsize
% \setlength{\tabcolsep}{3.0pt}
% \begin{tabular}{lccc}
% \toprule
% Method & Geom. RMSE & Phase RMSE & Cov. MAE \\
% \midrule
% Ours
% & $\mathbf{0.0210\pm0.0017}$
% & $\mathbf{0.0282\pm0.0027}$
% & $\mathbf{0.0347\pm0.0049}$ \\
% One-shot
% & $0.0541\pm0.0019$
% & $0.0687\pm0.0017$
% & $0.0813\pm0.0083$ \\
% Fixed coverage
% & $0.0571\pm0.0075$
% & $0.1036\pm0.0013$
% & $0.2070\pm0.0000$ \\
% Linear phase
% & $0.0945\pm0.0100$
% & $0.1431\pm0.0042$
% & $0.1359\pm0.0076$ \\
% \bottomrule
% \end{tabular}
% \end{table}

\begin{figure}[t]
    \centering

    \includegraphics[width=0.85\columnwidth]{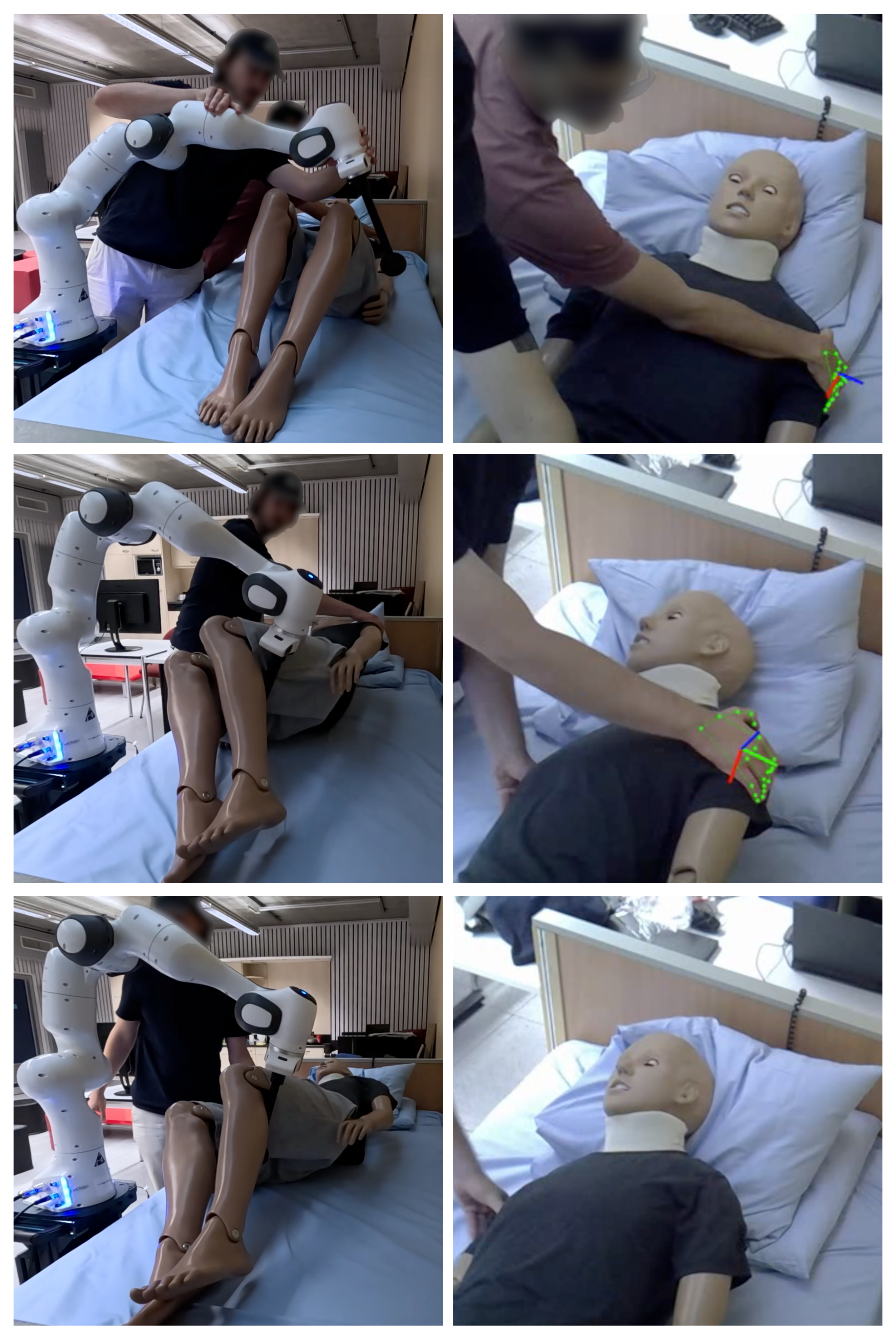}

    \caption{Patient-turning robot assistance experiment.
    From top to bottom: demonstration collection through guidance; nominal execution under continuous hand tracking; and online
    adaptation during temporary tracking loss. Figures on the right show camera
    observations used for phase estimation. During tracking loss, the current
    phase is retained until visual observation is recovered.}
    \label{fig:assistive_turning}
\end{figure}

\subsection{Robot Experiments}
\label{sec:real_robot}
As a downstream application of the proposed spacetime encoding and optimization approach, we demonstrate that the recovered phase can serve as an online execution
coordinate in a collaborative patient-turning task using a 7-DoF Franka
robot and a manikin. The demonstrated assistive task requires two people: one to move
the upper body and one to guide the lower body. During demonstration
collection, the second person kinesthetically guides the robot to dmeosntrate the synchronized assistive behavior. The resulting demonstration
establishes the correspondence between human motion progress and robot
motion. An RGB camera tracks the human hand at approximately
$15\,\mathrm{Hz}$.

At runtime, each valid hand observation is locally matched to the
demonstrated hand trajectory to estimate the human phase $\hat{s}_H$.
Before execution, the human hand trajectory is aligned with the demonstration
using a predefined shoulder pose. The estimated phase is constrained to
remain monotone, preventing temporal regression under visual noise or
ambiguous local correspondences. The human phase provides the reference
for the robot phase $s_R$, and the desired joint-space commands are
\begin{equation}
    \bm{q}_d=\bm{g}_R(s_R),
    \qquad
    \dot{\bm{q}}_d=\bm{g}'_R(s_R)\dot{s}_R,
\end{equation}
with commands updated at approximately $100\,\mathrm{Hz}$.
Robot progression is then determined by the observed human motion.

Fig.~\ref{fig:assistive_turning} shows nominal execution and online
adaptation using the same learned spacetime representation. Under continuous
tracking, the robot progresses synchronously with the human partner. In the
adaptation trial, the hand observation is temporarily removed during
execution, causing the phase to be held and the robot to maintain its current
configuration. Execution is finally resumed once hand tracking is recovered, demonstrating adaptation capabilities in a real world scenario (see also accompanying video). % The experiment demonstrates that temporary variations in visual availability can be handled online.

\section{Conclusion}
We presented a spacetime spline encoding strategy for learning robot
motions from complete and partial demonstrations.
The formulation jointly estimates a shared geometric path,
demonstration-specific temporal profiles, and coverage
intervals within a single augmented state space, while allowing
for coordinate-specific continuity constraints.
Simulated experiments demonstrate that the spacetime representation allows robust recovery under variations in the demonstration set, and adaptation to varying continuity in demonstrated paths. % We have further shown that the method can be parallelized for increased computational efficiency.
As an example use case, we considered an assistive robotics experiment requiring online retiming, as an illustration of adaptation capability in human-robot collaborative tasks requiring robustness to visual tracking loss. we plan in future work to further leverage the modularity and interpretability of the spacetime spline representation in various collaborative tasks involving multiple users and robots (incl. dual-arm and multi-finger tasks). 

\bibliographystyle{IEEEtran}
\bibliography{references}

\end{document}